\documentclass[12pt]{article}

\usepackage[utf8]{inputenc}
\usepackage[margin=1in]{geometry}
\usepackage{booktabs}
\usepackage{amsmath,amssymb}
\usepackage{graphicx}
\usepackage{listings}
\usepackage{enumitem}
\usepackage{tikz}
\usepackage{xcolor}
\usepackage{setspace}
\usepackage{hyperref}
\usepackage[natbibapa]{apacite}

\usetikzlibrary{shapes.geometric, arrows.meta, positioning, fit, calc}

\begin{document}
\doublespacing

\begin{center}
    \Large\textbf{Slide Deck Q\&A Quality Assurance App: A Multi-Stage Pipeline for Pedagogical Question Generation}
\end{center}

\vspace{3em}

\begin{center}
    \textbf{Jim Salsman} \\
    TalkNicer, Inc. \\
    jim@talknicer.com \\
    \url{https://orcid.org/0000-0002-1715-7990}
\end{center}

\vspace{3em}

\begin{abstract}
Generating high-quality, pedagogically useful questions from lecture slide decks is difficult because important instructional content is distributed across both text and visual elements, and because useful questions must be scaffolded across the flow of a presentation rather than generated slide by slide in isolation. This paper describes Slide Deck Q\&A Quality Assurance (slidesqaqa), a Flask-based software system that extracts text and rendered images from PDF slides and processes them through a four-stage large language model pipeline comprising window planning, deck synthesis, slide annotation, and reconciliation. The system reasons jointly about slide modality and pedagogical role, allocates bounded question budgets, and revises draft annotations at the deck level to reduce redundancy and improve coverage. The final output is a structured JSON annotation containing deck-level goals, section structure, slide-level summaries, question sets, and evaluation scores. Initial experiments on two technical lecture decks indicate that the pipeline can filter non-instructional slides and produce high-fidelity, pedagogically coherent questions for visually complex content.
\\ \\
The working system is at \url{https://slidesqaqa-974767694043.us-west1.run.app}
\\ \\
The software repository is at \url{https://github.com/blinding2submit/slidesqaqa}
\end{abstract}

\vspace{1em}
\vspace{2em}

\section*{Introduction}

Educators frequently rely on presentation slide decks as the primary medium for delivering instructional content, yet creating targeted, high-fidelity comprehension questions for each slide requires significant manual effort and pedagogical expertise. While automated approaches to question generation have been explored to alleviate this burden, they often struggle with the inherent multi-modal nature of modern presentations, where information is distributed across text, images, and diagrams, and limited by the context windows of underlying models. Developing robust automated systems is critical because high-quality pedagogical questions are known to significantly enhance active learning, student engagement, and long-term retention. Standard generative text models, which form the basis of many early automated tools, typically struggle to interpret the visual information—such as charts, graphs, or complex diagrams—that carries much of the semantic weight in contemporary slide decks.

To address these challenges, we present slidesqaqa, a novel software system designed to automate the extraction and generation process. slidesqaqa takes a PDF presentation as input, renders each individual slide into a visual thumbnail, extracts the corresponding raw text, and meticulously coordinates a sophisticated four-phase Large Language Model (LLM) workflow comprising Window Planning, Deck Synthesis, Slide Annotation, and Reconciliation. The system ultimately outputs a comprehensive, hierarchical JSON structure that details the deck's topics, categorizes slide roles, defines specific learning goals, and provides bounded, context-aware comprehension questions.

The main contributions of this paper are threefold: first, we introduce a windowed deck-planner approach designed specifically to mitigate LLM context limits while preserving the critical narrative flow of the presentation; second, we propose static heuristics for dynamically balancing question budgets based on the detected modality of each slide; and third, we detail a multi-pass generative architecture that culminates in a dedicated reconciliation step to ensure consistency and quality. The remainder of this paper is structured as follows: we first review related work in automated question generation and multimodal analysis, then outline the system's goals and high-level architecture, before detailing its concrete implementation using Flask and PyMuPDF and discussing the methods used for inference. Finally, we conclude with a discussion of future avenues for evaluation and acknowledging current system limitations.

\subsection*{Related Work}
Traditional automated question generation (AQG) has heavily focused on text-based inputs using NLP techniques to convert declarative sentences into educational questions, establishing foundational metrics for evaluating coverage and pedagogical utility \citep{kurdi2020systematic,gorgun2024exploring}. However, modern pedagogical materials, particularly slide decks, are inherently multimodal.

The task of extracting information from these visual and textual modalities closely relates to Visual Question Answering (VQA) \citep{kim2025vqa}. The recent emergence of multimodal large language models (MLLMs) provides a strong baseline architecture capable of processing interleaved text and image inputs \citep{yin2024multimodal}. Yet, applying standard MLLMs or naive text-only pipelines to slide decks often leads to poor performance. Research has shown that Vision-Language Models (VLMs) can struggle with fine-grained evidence grounding and modality reliance, especially when interpreting complex layouts involving diagrams, charts, and formulas \citep{sim2025vlms}.

Furthermore, existing systems often process documents in single large chunks or disjointed pages, ignoring structural flow and relying on brittle OCR. This approach loses narrative coherence and fails to capture layout-dependent semantics crucial for effective visually rich document retrieval \citep{zhang2025roles}. slidesqaqa addresses these gaps by stitching rendered thumbnails into multi-slide contact sheets using a sliding-window chunking strategy to retain transitional context, and by implementing a dynamic reconciliation phase to balance question budgets across heterogeneous slide types.

\subsection*{Research Questions}
This study seeks to answer the following research questions regarding the automated generation of educational questions from multimodal slide decks:

\begin{itemize}[leftmargin=*]
    \item \textbf{RQ1:} To what extent does a multi-stage overlapping window pipeline improve the pedagogical scaffolding of generated questions compared to naive per-slide generation?
    \item \textbf{RQ2:} How does explicit consideration of slide modality and role affect the fidelity (groundedness) and coverage of the resulting question sets?
    \item \textbf{RQ3:} Can static heuristics combined with a dynamic reconciliation phase effectively balance question budgets across heterogeneous slide types without manual intervention?
\end{itemize}

\section*{Method}
\subsection*{System Overview}

The high-level architecture of slidesqaqa is structured as a robust, single-server Python Flask application, which serves as the central hub coordinating both document processing and complex model interactions. At its core, the system relies on PyMuPDF for high-fidelity document extraction, meticulously pulling out structural components and textual content from uploaded presentations. For the generation of semantic insights and pedagogical questions, the application integrates tightly with the Google GenAI SDK, providing the necessary inference capabilities to process multimodal inputs effectively.

This architecture is composed of three main, tightly coupled components, each fulfilling a critical role in the pipeline. The \textit{Preprocessor} acts as the initial ingestion layer; it directly reads the uploaded PDF document, systematically extracts all available native text, and simultaneously renders high-resolution PNG images of each slide to capture its visual layout. Following extraction, the \textit{LLM Pipeline Engine} takes over, serving as the system's orchestrator. It manages the execution of four distinct, sequential generation phases—Window Planning, Deck Synthesis, Slide Annotation, and Reconciliation—while strictly enforcing that all outputs conform to strongly typed JSON schemas, ensuring data integrity. Finally, the \textit{Frontend UI} provides the user interface; it is an embedded HTML and JavaScript page designed to function as both a real-time log viewer, providing transparency into the generation process, and a dynamic data visualizer for the final output.

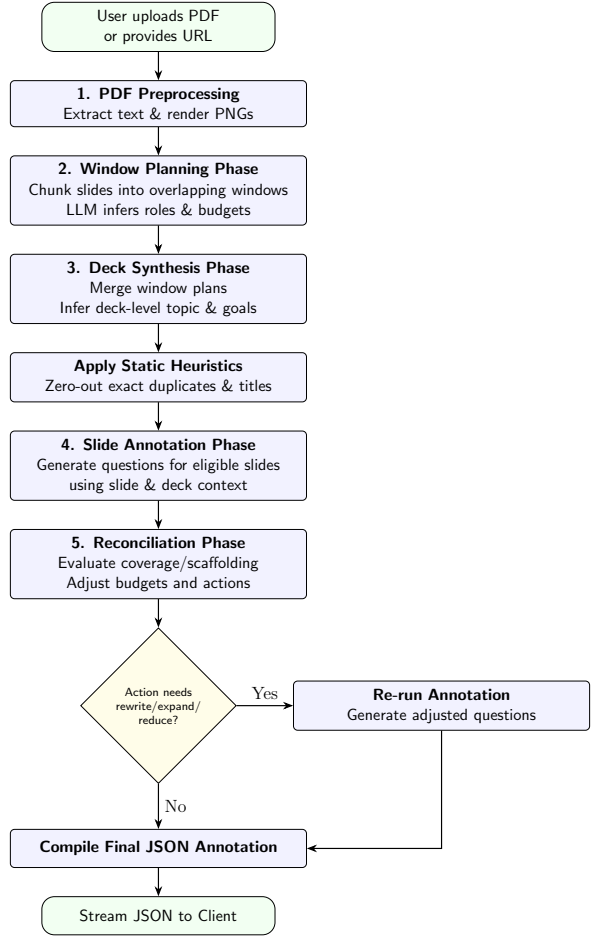
\begin{figure}[!t]
    \centering
    \resizebox{!}{0.54\textheight}{%
    \begin{tikzpicture}[
        node distance=0.68cm and 0.95cm,
        box/.style={rectangle, draw, rounded corners=1mm, text centered, text width=6.7cm, minimum height=2.2em, inner sep=5pt, font=\sffamily\small, fill=blue!5},
        rounded/.style={rectangle, draw, rounded corners=3mm, text centered, text width=5.2cm, minimum height=2.2em, inner sep=5pt, font=\sffamily\small, fill=green!5},
        decision/.style={diamond, draw, text centered, text width=2.7cm, inner sep=1pt, font=\sffamily\scriptsize, fill=yellow!10},
        arrow/.style={-{Stealth[scale=1.05]}, thick}
    ]

        \node[rounded] (start) {User uploads PDF or provides URL};
        \node[box, below=of start] (preprocess) {\textbf{1. PDF Preprocessing}\\Extract text \& render PNGs};
        \node[box, below=of preprocess] (windowplan) {\textbf{2. Window Planning Phase}\\Chunk slides into overlapping windows\\LLM infers roles \& budgets};
        \node[box, below=of windowplan] (synth) {\textbf{3. Deck Synthesis Phase}\\Merge window plans\\Infer deck-level topic \& goals};
        \node[box, below=of synth] (heuristics) {\textbf{Apply Static Heuristics}\\Zero-out exact duplicates \& titles};
        \node[box, below=of heuristics] (annotate) {\textbf{4. Slide Annotation Phase}\\Generate questions for eligible slides\\using slide \& deck context};
        \node[box, below=of annotate] (recon) {\textbf{5. Reconciliation Phase}\\Evaluate coverage/scaffolding\\Adjust budgets and actions};
        \node[decision, below=of recon] (rerun) {Action needs\\rewrite/expand/\\reduce?};

        \node[box, right=1.35cm of rerun] (reannotate) {\textbf{Re-run Annotation}\\Generate adjusted questions};
        \node[box, below=1.08cm of rerun] (compile) {\textbf{Compile Final JSON Annotation}};
        \node[rounded, below=of compile] (end) {Stream JSON to Client};

        \draw[arrow] (start) -- (preprocess);
        \draw[arrow] (preprocess) -- (windowplan);
        \draw[arrow] (windowplan) -- (synth);
        \draw[arrow] (synth) -- (heuristics);
        \draw[arrow] (heuristics) -- (annotate);
        \draw[arrow] (annotate) -- (recon);
        \draw[arrow] (recon) -- (rerun);
        \draw[arrow] (rerun) -- node[above] {Yes} (reannotate);
        \draw[arrow] (rerun) -- node[right] {No} (compile);
        \draw[arrow] (reannotate) |- (compile);
        \draw[arrow] (compile) -- (end);

    \end{tikzpicture}%
    }
    \caption{Conceptual LLM pipeline data flow for the slidesqaqa application.}
    \label{fig:architecture}
\end{figure}

The typical data flow through the slidesqaqa system (as illustrated in Figure \ref{fig:architecture}) is linear and highly structured. The process initiates with a user uploading a PDF document, which immediately triggers the Text and Image extraction phase handled by the Preprocessor. The extracted data is then fed into the Window Planning stage, where initial contextual windows are formed. This is followed by Synthesis and Heuristics, where the overall deck structure is inferred and question budgets are allocated. The pipeline then proceeds to Slide Annotation, where the actual pedagogical questions are generated based on the prior analysis. The Reconciliation phase subsequently reviews and refines these annotations to ensure consistency and adherence to the allocated budgets. Ultimately, the refined data undergoes Final JSON Compilation, assembling the results into a unified structure that is seamlessly streamed back to the Client via the Frontend UI.

\subsection*{Design and Architecture}
The preprocessor relies on the `fitz` library (PyMuPDF) to extract native text and render scaled PNG images representing the visual layout of each slide.

The LLM Generation Pipeline conducts four major passes:
1. \textbf{Window Planner:} Analyzes overlapping windows of slides, creating a contact sheet image to infer titles, summaries, modalities, roles, and a provisional question budget.
2. \textbf{Deck Synthesis:} Merges window analyses into one coherent deck plan.
3. \textbf{Slide Annotator:} Analyzes individual slides requiring questions based on the deck plan, generating specific questions.
4. \textbf{Reconciliation:} Evaluates the provisional full-deck annotation for redundancies and unbalanced coverage.

The core data structures are strictly typed using Pydantic models (\texttt{SlidePlan}, \texttt{SlideAnnotationModel}, \texttt{ReconciliationModel}). The backend relies on Flask with a streaming endpoint, and the frontend uses vanilla JavaScript.

\subsection*{Experimental Setup}
To evaluate the system, we analyzed academic lecture slide decks focusing on complex subjects such as Natural Language Processing and Deep Learning. Specifically, we used two comprehensive decks: ``Self-Attention and Transformers'' (\texttt{cs224n-2024-lecture08-transformers})\footnote{Data shown at \url{https://pastebin.com/raw/5G0s2r2p}.} and ``Neural Constituency Parsing'' (\texttt{cs288-sp23-neural-parsing})\footnote{Data shown at \url{https://pastebin.com/raw/m15KdeVG}.}. These decks were selected because they prominently feature structural slides (e.g., titles, agendas) as well as highly complex, mechanism-heavy slides relying on intricate diagrams (e.g., LSTM bidirectional routing, mean-pooling aggregation, and span classification).

We evaluate the generated question sets across three key metrics derived from the multi-stage pipeline logs: Coverage (alignment with slide content and learning goals), Fidelity (accuracy and grounding in both visual and textual evidence), and Scaffolding (logical progression and pedagogical relevance of the questions in the context of the entire deck). The system automatically evaluates these metrics on a 1-5 scale during the generation process, as summarized in Table \ref{tab:qualitative_results}.

\begin{table}[htbp]
\centering
\caption{Qualitative summary of experimental results across evaluated slide decks.}
\label{tab:qualitative_results}
\begin{tabular}{@{}llp{6.5cm}@{}}
\toprule
\textbf{Deck} & \textbf{Metric Focus} & \textbf{Observation} \\ \midrule
Self-Attention and Transformers & Filtering & Correctly zero-budgeted administrative and transition slides. \\
Neural Constituency Parsing & Fidelity & Correctly generated questions addressing bidirectional routing, citing ``horizontal arrows pointing left and right between the boxes''. \\
Self-Attention and Transformers & Coverage & Generated a question asking how ``Sentence encoding'' is computed, identifying the answer as ``taking the element-wise max or mean of all hidden states''. \\
Both Decks & Scaffolding & Scaffolding scores remained high (3--5); generated items served as solid foundational checks and progressed logically. \\ \bottomrule
\end{tabular}
\end{table}

\section*{Results}
The multi-stage pipeline reliably filtered structural slides and concentrated generation efforts on core pedagogical material. For instance, in the ``Self-Attention and Transformers'' deck, the reconciliation phase correctly zero-budgeted administrative and transition slides, reducing the overall question load while maintaining narrative coherence. When generating questions, the system achieved high quantitative marks: questions consistently earned Fidelity scores of 5 and Coverage scores of 4 to 5 across both lecture decks. Scaffolding scores also remained high (typically 3 to 5), indicating that the generated items served as solid foundational checks and progressed logically.

Qualitatively, the slidesqaqa pipeline successfully interpreted complex multimodal diagrams. For example, in a slide detailing mean-pooling as a conceptual stepping stone toward attention, the system generated a question asking how ``Sentence encoding'' is computed, correctly identifying the answer as ``taking the element-wise max or mean of all hidden states'' (Fidelity: 5, Coverage: 5). Similarly, in the ``Neural Constituency Parsing'' deck, when interpreting a visual diagram of LSTM units, the system correctly generated questions addressing how bidirectional routing works, accurately citing ``horizontal arrows pointing left and right between the boxes'' as evidence for how context is gathered across the sequence (Fidelity: 5).

\section*{Discussion}
The experimental results strongly support the hypothesis that a multi-stage, layout-aware pipeline is necessary for generating high-quality pedagogical questions from slide decks. The high fidelity and coverage scores demonstrate the value of the Window Planner and Reconciliation phases. slidesqaqa's ability to identify and reduce question budgets on redundant or transitional slides directly addresses a major limitation in existing naive AQG tools that treat every slide equally.

The primary strength of this system is its robust handling of multi-modal information. The system was able to successfully identify visual cues, such as ``horizontal arrows,'' and map them to complex architectural concepts like bidirectional context gathering in neural networks. However, we acknowledge several limitations. Processing large decks incurs substantial LLM inference latency and API costs due to the multi-pass architecture. Furthermore, the reliance on a specific proprietary LLM (Gemini) poses potential risks regarding reproducibility and long-term stability if the underlying API models are updated or deprecated.

\section*{Conclusion}
slidesqaqa demonstrates a sophisticated approach to automated multi-modal question generation, structuring instructional content extraction into a resilient four-stage pipeline. Chunking slides into visual contact sheets and rigorously employing type-validated LLM generation significantly enhances the quality of structural and pedagogical analysis. Future work may include comprehensive evaluation against human-authored question sets, latency optimizations, and exploring fine-tuned, smaller models for the window planning phases.

\section*{Disclosures}

No external funding was received for this study. The author declares no competing interests. The data used in this study consist of lecture slide decks. The datasets supporting the conclusions of this article are available at \url{https://pastebin.com/raw/5G0s2r2p} and \url{https://pastebin.com/raw/m15KdeVG}. The author was responsible for conceptualization, methodology, software, investigation, formal analysis, writing all drafts, review, and editing. During the preparation of this work the author used Google Gemini Pro 3.1, Google Jules, and OpenAI ChatGPT 5.4 in order to produce both code and prose. After using those tools, the author reviewed and edited the content thoroughly and takes full responsibility for the app and publication content.

\appendix
\section*{Appendix A: Output JSON Schema}
This appendix describes the structure and descriptions of the final output JSON schema generated by the slidesqaqa application.

\begin{lstlisting}[basicstyle=\ttfamily\tiny, breaklines=true]
{
  "schema_version": "Version of this JSON schema.",
  "field_descriptions": {
    "...": "Field descriptions mapping"
  },
  "deck_metadata": {
    "deck_id": "Stable identifier for this deck.",
    "deck": "Full academic citation for the deck.",
    "deck_url": "Original source URL for the deck, if known.",
    "source_file": "Local uploaded PDF filename.",
    "total_slides": "Total number of PDF pages processed as slides.",
    "processed_at": "UTC timestamp when this JSON was produced."
  },
  "deck_analysis": {
    "deck_topic": "Short description of the overall topic of the deck.",
    "target_audience": "Estimated audience level; for example undergraduate, graduate, or mixed.",
    "learning_goals": ["List of deck-level learning goals inferred from the slides."],
    "sections": [
      {
        "section_id": "String",
        "start_slide": "Integer",
        "end_slide": "Integer",
        "section_title": "String",
        "section_summary": "String"
      }
    ],
    "coverage_targets": ["Deck-level content targets such as text, diagram, table, chart, layout-aware, or image-plus-text."],
    "global_notes": "Important global caveats, ambiguities, or observations."
  },
  "reconciliation": {
    "revised_slide_actions": [
      {
        "slide_number": "Integer",
        "action": "String",
        "new_question_budget": "Integer",
        "reason": "String"
      }
    ],
    "deck_reconciliation_notes": "Global notes about redundancy, balancing, and quality adjustments across the deck.",
    "uncovered_learning_goals": ["Deck learning goals that remain weakly covered after reconciliation."],
    "redundancy_warnings": ["Warnings about overlapping or repeated question sets across slides."]
  },
  "slides": [
    {
      "slide_id": "Stable identifier for a slide within the deck.",
      "slide_number": "1-based slide number corresponding to the PDF page order.",
      "slide_title": "Visible title on the slide if present; otherwise a concise generated title.",
      "modality_type": "Dominant visual form of the slide; for example text, diagram, table, chart, layout-aware, image-plus-text, or mixed.",
      "role_in_deck": "Instructional role of the slide within the deck; for example title, agenda, transition, definition, example, mechanism, result, summary, or appendix.",
      "local_summary": "One- or two-sentence summary of the slide's main instructional content.",
      "key_concepts": ["List of key concepts explicitly present on the slide."],
      "evidence_regions": ["List of human-readable descriptions of important visible regions on the slide."],
      "eligible_for_questions": "Whether the slide should receive any comprehension questions.",
      "eligibility_reason": "Explanation for why the slide should or should not receive questions.",
      "question_budget": "Recommended number of questions for this slide in deck context.",
      "question_mix": ["Recommended mix of question types for this slide."],
      "questions": [
        {
          "question_id": "Stable identifier for a question within a slide.",
          "question_type": "Controlled label for the question form or reasoning type.",
          "prompt": "Question text shown to the learner.",
          "options": ["List of answer options for a multiple-choice item; empty otherwise."],
          "answer": "Gold answer or bounded reference answer grounded in the slide.",
          "evidence_span": "Short description of where the answer is visible on the slide.",
          "difficulty": "Relative difficulty label such as low, medium, or high.",
          "purpose": "Instructional purpose such as terminology, relation check, interpretation, or synthesis.",
          "fidelity_score": "1-5 judgment of whether the question is answerable from the slide alone.",
          "fidelity_notes": "Short rationale for the fidelity score."
        }
      ],
      "evaluation": {
        "coverage_score": "1-5 score for how well the slide's question bundle covers the slide's important content; null when the slide intentionally has no questions.",
        "coverage_notes": "Short rationale for the coverage score.",
        "scaffolding_score": "1-5 score for how well the question bundle forms an instructional progression; null when the slide intentionally has no questions.",
        "scaffolding_notes": "Short rationale for the scaffolding score."
      }
    }
  ]
}
\end{lstlisting}

\section*{Appendix B: LLM Prompts}
This appendix contains the exact system prompt strings used in the four-phase pipeline of the slidesqaqa application.

\begin{lstlisting}[basicstyle=\ttfamily\tiny, breaklines=true]
WINDOW_PLANNER_PROMPT = """
You are analyzing a contiguous window from a larger lecture slide deck.

For each slide in this window:
- infer slide_title
- write a short local_summary
- assign modality_type
- assign role_in_deck
- decide whether the slide is eligible for learner-facing comprehension questions
- give an eligibility_reason
- assign a question_budget from 0 to 5
- assign a question_mix

Important:
- Use neighboring slides in the window to reason about redundancy and transitions.
- It is acceptable to assign zero questions.
- Do not force a fixed number of questions.
- Favor low budgets for title, agenda, transition, administrative, appendix, and repeated recap slides.
- Favor higher budgets for rich mechanism, comparison, result, diagram, chart, table, or synthesis slides.
- question_mix must use only these values:
  ["fill_blank", "mcq", "open_ended", "short_answer", "diagram_labeling", "comparison", "interpretation", "evidence_localization"]
- modality_type must use only these values:
  ["text", "diagram", "table", "chart", "layout-aware", "image-plus-text", "mixed"]
- role_in_deck must use only these values:
  ["title", "agenda", "transition", "definition", "example", "mechanism", "comparison", "result", "summary", "administrative", "appendix", "review", "reference"]

Return JSON only. Do not include explanatory prose outside JSON.
"""

DECK_SYNTHESIS_PROMPT = """
You are merging overlapping window-level analyses of one lecture slide deck into one final deck plan.

Goals:
1. Infer the deck topic and likely target audience.
2. Infer deck-level learning goals.
3. Produce section boundaries for the full deck.
4. Resolve conflicting window-level slide plans conservatively.
5. Return exactly one slide plan object per slide number.

Important:
- Preserve zero-question slides when they are non-instructional, redundant, or too thin.
- Some slides may deserve more than three questions.
- Keep question budgets based on instructional importance, self-containedness, evidence richness, and novelty.
- Use only the allowed label vocabularies already present in the window plans.
- Sections should be contiguous and ordered.

Return JSON only. Do not include explanatory prose outside JSON.
"""

SLIDE_ANNOTATOR_PROMPT = """
You are generating slidesqaqa annotations for one slide within a lecture deck.

Use both the local slide evidence and the provided deck context.

Your tasks:
1. Identify key_concepts explicitly present on the slide.
2. Identify 2 to 6 evidence_regions as short human-readable descriptions of important visible regions.
3. Generate exactly the assigned question budget in the supplied question mix.
4. Every question must be answerable from the slide alone.
5. Every answer must be bounded and evidence-grounded.
6. Use deck context only to decide what is educationally important. Do not answer from hidden lecture knowledge.
7. Avoid redundancy with the neighboring slides when possible.

Question-writing guidance:
- On text slides, favor terminology, distinctions, and concise explanation.
- On diagram slides, favor component labeling, relationships, flow, and mechanism.
- On table/chart slides, favor lookup, comparison, trend, and interpretation.
- On layout-aware slides, favor spatial or grouping-based reasoning when relevant.
- If a question_type is mcq, include exactly 4 options.
- If a question_type is not mcq, options must be an empty list.
- fidelity_score must be an integer from 1 to 5.
- coverage_score and scaffolding_score must be integers from 1 to 5.

Coverage guidance:
- 1 means poor coverage or repeated tiny facts.
- 3 means adequate coverage of the main concept and at least one secondary element.
- 5 means strong coverage of the slide's important visible content.

Scaffolding guidance:
- 1 means random or disconnected.
- 3 means reasonable progression.
- 5 means coherent progression from simpler to deeper understanding.

Return JSON only. Do not include explanatory prose outside JSON.
"""

RECONCILIATION_PROMPT = """
You are reconciling a provisional slidesqaqa annotation set for a full lecture deck.

You are given:
- deck metadata
- deck analysis
- all slide plans
- all provisional slide annotations

Your task is to improve the deck as a whole.

Goals:
1. Detect redundant question sets across nearby slides.
2. Detect slides that should have fewer questions.
3. Detect rich slides that deserve more questions.
4. Detect places where learning-goal coverage is unbalanced.
5. Detect weak scaffolding within sections.

Rules:
- Do not force similar budgets across all slides.
- Preserve zero-question slides when they are truly non-instructional or redundant.
- Prefer deleting weak or redundant questions rather than inventing extra ones.
- Use this action vocabulary only:
  ["keep", "reduce", "expand", "zero_out", "rewrite"]
- For each slide, return one action and a new_question_budget between 0 and 5.
"""
\end{lstlisting}

\end{document}